# From ACR O-RADS 2022 to Explainable Deep Learning: Comparative Performance of Expert Radiologists, Convolutional Neural Networks, Vision Transformers, and Fusion Models in Ovarian Masses

Ali Abbasian Ardakani[1]; Afshin Mohammadi[2]; Alisa Mohebbi[1]; Anushya Vijayananthan[3]; Sook Sam Leong[4]; Lim Yi Ting[3]; Mohd Kamil Bin Mohamad Fabell[3]; U Rajendra Acharya[5,6]; Sepideh Hatamikia[1,7,8*]

[1] *Clinical AI-Research in Omics and Medical Data Science (CAROM) group, Department of Medicine, Faculty of Medicine and Dentistry, Danube Private University, Krems an der Donau, Austria*
[2] *Department of Radiology, Faculty of Medicine, Urmia University of Medical Science, Urmia, Iran*
[3] *Department of Biomedical Imaging, Universiti Malaya Research Imaging Centre, Faculty of Medicine, Universiti Malaya, Kuala Lumpur, Malaysia*
[4] *Centre for Medical Imaging Studies, Faculty of Health Sciences, Universiti Teknologi MARA Selangor, Selangor, Malaysia*
[5] *School of Mathematics, Physics and Computing, University of Southern Queensland, Springfield, Queensland, Australia*
[6] *Centre for Health Research, University of Southern Queensland, Springfield, Queensland, Australia*
[7] *Austrian Center for Medical Innovation and Technology (ACMIT), Wiener Neustadt, Austria*
[8] *Center for Medical Physics and Biomedical Engineering, Medical University of Vienna*

∗ **Corresponding author:**
**Dr. Sepideh Hatamikia**
**E-mail:** Sepideh.hatamikia@DP-Uni.ac.at, **ORCID:** https://orcid.org/0000-0002-0182-0954

# From ACR O-RADS 2022 to Deep Learning: Comparative Performance of Expert Radiologists, Convolutional Neural Networks, Vision Transformers, and Fusion Models in Adnexal Masses

## Abstract:

**Background:** The 2022 update of the Ovarian-Adnexal Reporting and Data System (O-RADS) ultrasound classification refines risk stratification for adnexal lesions, yet human interpretation remains subject to variability and conservative thresholds. Concurrently, deep learning (DL) models have demonstrated promise in image-based ovarian lesion characterization. This study evaluates radiologist performance applying O-RADS v2022, compares it to leading convolutional neural network (CNN) and Vision Transformer (ViT) models, and investigates the diagnostic gains achieved by hybrid human-AI frameworks.

**Methods:** In this single-center, retrospective cohort study, a total of 512 adnexal mass images from 227 patients (110 with at least one malignant cyst) were included. Sixteen DL models, including DenseNets, EfficientNets, ResNets, VGGs, Xception, and ViTs, were trained and validated. A hybrid model integrating radiologist O-RADS scores with DL-predicted probabilities was also built for each scheme.

**Results:** Radiologist-only O-RADS assessment achieved an AUC of 0.683 and an overall accuracy of 68.0%. CNN models yielded AUCs of 0.620 to 0.908 and accuracies of 59.2% to 86.4%, while ViT16-384 reached the best performance, with an AUC of 0.941 and an accuracy of 87.4%. Hybrid human-AI frameworks further significantly enhanced the performance of CNN models; however, the improvement for ViT models was not statistically significant (P-value >0.05).

**Conclusions:** DL models markedly outperform radiologist-only O-RADS v2022 assessment, and the integration of expert scores with AI yields the highest diagnostic accuracy and discrimination. Hybrid human-AI paradigms hold substantial potential to standardize pelvic ultrasound interpretation, reduce false positives, and improve detection of high-risk lesions.

**Keywords:** O-RADS v2022, Ovarian Cancer, Vision Transformer, Ultrasound, Deep Learning

## 1. Introduction

The Ovarian-Adnexal Reporting and Data System (O-RADS) ultrasound was first introduced in 2018 by the American College of Radiology (ACR) to provide a standardized lexicon for describing ovarian and adnexal findings, subsequently evolving through iterations in 2020 and ultimately culminating in the comprehensive 2022 version. The ACR O-RADS 2022 update represents a significant advancement in risk stratification, incorporating additional morphologic features that favor benignity, including bilocular characteristics for cysts without solid components and shadowing features for solid lesions with smooth contours, while improving diagnostic specificity compared to its predecessors [1, 2].

Parallel to these advances in standardized reporting systems, artificial intelligence (AI) has emerged as a transformative technology in medical imaging. Deep learning (DL) approaches, encompassing convolutional neural networks (CNNs) and vision transformers (ViTs), have demonstrated remarkable potential in automated ovarian mass characterization using ultrasound images. Contemporary investigations have explored diverse architectures, including DenseNet, ResNet, EfficientNet, and ViT families [3-7]. The comparative performance of AI models against established reporting systems has yielded compelling results, with several investigations demonstrating that DL algorithms can achieve diagnostic performance comparable to ACR O-RADS assessments and expert subjective evaluations. Gao and colleagues reported that their CNN model achieved diagnostic accuracy matching expert radiologists while improving the performance of less experienced examiners when used as an assistive tool [8]. Similarly, Chen et al. demonstrated that feature fusion DL models achieved AUCs of 0.93, comparable to ACR O-RADS (AUC 0.92) and expert assessment (AUC 0.97) [7].

The synergistic combination of AI models with radiologist assessment represents an emerging paradigm that leverages the complementary strengths of both approaches. Ensemble models integrating radiomics features, DL characteristics, and clinical data have shown superior performance compared to individual modalities, with studies reporting improvements in diagnostic accuracy when AI assistance is provided to radiologists [9]. The additive value of combining AI with standardized reporting systems has been particularly pronounced in challenging cases, such as ACR O-RADS category 4 lesions, where diagnostic uncertainty remains highest [10]. Despite these advances, a critical gap exists in the literature regarding the comparative evaluation of AI models against the most recent ACR O-RADS 2022 version. Previous studies have predominantly compared or added their AI algorithms with ACR O-RADS 2018 and 2020 versions, while the enhanced specificity and refined risk stratification capabilities of ACR O-RADS 2022 remain unexplored in the context of AI-assisted diagnosis [5, 11, 12]. Furthermore, the potential additive value of combining multiple AI architectures with ACR O-RADS 2022 assessments represents an understudied area that could significantly impact clinical decision-making and patient outcomes.

The present investigation addresses these knowledge gaps by conducting the first comprehensive comparison of radiologist performance using ACR O-RADS 2022 versus multiple state-of-the-art AI models, including both CNN architectures (DenseNets, ResNets, EfficientNets, VGGs, and Xception) and ViT variants. Moreover, this study uniquely evaluates the additive diagnostic value of combining AI models with ACR O-RADS 2022 assessments, representing the updated contemporary evaluation of hybrid human-AI diagnostic approaches in ovarian imaging.

## 2. Methods

This retrospective single-center comparative study was conducted following the principles of the Declaration of Helsinki and was approved by the institutional review board. The study was designed as a comprehensive evaluation comparing the diagnostic performance of human radiologists using the ACR O-RADS 2022 classification system [1] against various AI models, as well as the synergistic performance when both approaches are combined.

### *2.1. Patient Selection & Study Design*

The study encompassed patients from a tertiary care center specializing in gynecological oncology and reproductive medicine. The study population included consecutive patients who underwent transvaginal ultrasound examination for suspected ovarian or adnexal masses between January 2016 and May 2017. Patients were identified through review of electronic health records and picture archiving and communication systems.

The inclusion criteria were as follows: (A) patients aged 18 years or older with suspected ovarian or adnexal masses identified on ultrasound imaging; (B) availability of high-quality B-mode ultrasound images; (C) confirmation through histopathology (surgical excision within one month of imaging) or adequate follow-up (for at least 12 months for conservatively managed lesions); (D) fall within O-RADS risk categories 2 through 5 to focus on masses where diagnostic uncertainty is greatest, excluding purely normal ovaries or unequivocally benign cysts whose minimal clinical impact could dilute performance comparisons.

Patients were excluded if they met any of the following criteria: (A) poor image quality that precluded adequate assessment; (B) patients with a history of ovarian malignancy, prior chemotherapy or radiotherapy involving the pelvis, or previous ovarian surgery were omitted to prevent pre-existing tissue changes or scarring from confounding image interpretation; (C) pregnant patients and those with active pelvic inflammatory disease, endometriosis flares, or adnexal torsion at the time of scan because they may mimic or mask neoplastic features and introduce diagnostic ambiguity unrelated to underlying tumor biology; (D) incomplete follow-up or pathological data; (E) presence of concurrent pelvic masses of extra-ovarian origin (e.g., leiomyomas, para-ovarian cysts, or gastrointestinal or urinary tract lesions); and (F) to avoid spectrum bias and ensure stability of lesion morphology, cases in which surgical specimens were obtained >1 month after imaging were excluded, as intervening growth or treatment could alter lesion characteristics.

Unlike most prior investigations that restricted analyses to a single cystic lesion per patient, this study, retained all eligible lesions regardless of count, thereby reflecting real-world clinical complexity and avoiding potential selection bias. Furthermore, whereas many studies have limited confirmation only to surgically resected specimens, our cohort was expanded to include conservatively managed patients with at least twelve months of clinical and imaging follow-up, ensuring preservation of true low-risk O-RADS classifications and preventing inadvertent exclusion of benign lesions that never underwent surgical intervention.

## *2.2. Pathological Classification & Longitudinal Follow-up*

The lesions underwent standardized histopathological evaluation by board-certified gynecologic pathologists, categorizing tumors into benign and malignant (including borderline subtype) groups while being blinded to O-RADS categorizations and clinical data. For lesions managed conservatively without surgery, patients received systematic clinical and sonographic surveillance at three- to six-month intervals for a minimum of twelve months, with any interval change in size or morphology triggering further diagnostic workup. Lesions that remained stable or regressed throughout follow-up were deemed benign, thereby preserving low O-RADS designations and avoiding misclassification bias. All pathological and follow-up data were independently reviewed and adjudicated by a multidisciplinary tumor board to ensure consistency and validity of the reference standards.

## 2.3. Ultrasound Image Acquisition & Protocol

All examinations were performed with high-resolution transvaginal and transabdominal probes using the Philips EPIQ ultrasound machine. Sonographers adhered to a uniform protocol specifying probe frequency, depth, focal zone, gain, and dynamic range settings to optimize lesion visualization. Gray-scale images in orthogonal planes captured maximal lesion dimensions and internal architecture. Immediately following each study, images were reviewed for technical adequacy; cases exhibiting motion artifacts, poor contrast resolution, or incomplete lesion coverage were reacquired where possible or excluded. All image data were archived in DICOM format and underwent centralized quality control by a panel of expert sonographers who verified labeling accuracy, adherence to imaging parameters, and absence of acquisition artifacts before inclusion in radiologist and AI analyses.

## 2.4. Radiologist Evaluation

The ACR O-RADS 2022 was implemented as the reference standard for radiologist assessment [1]. The O-RADS system stratifies ovarian lesions into six risk categories: O-RADS 0 representing incomplete evaluation due to technical limitations or patient factors; O-RADS 1 indicating normal premenopausal ovary with 0% malignancy risk; O-RADS 2 encompassing almost certainly benign lesions with less than 1% malignancy risk, including classic hemorrhagic cysts, typical endometriomas, mature cystic teratomas, and simple unilocular cysts; O-RADS 3 representing low-risk lesions with 1% to 10% malignancy risk, including unilocular cysts with irregular walls or larger simple cysts; O-RADS 4 indicating intermediate-risk lesions with 10% to 50% malignancy risk, including multilocular cystic lesions and unilocular-solid masses with papillary projections; and O-RADS 5 representing high-risk lesions with greater than 50% malignancy risk, including solid masses and lesions with ascites or peritoneal involvement. All ultrasound examinations and manual segmentations were performed by a single radiologist with 24 years of ultrasound interpretation experience. Reports were independently double-checked by a second radiologist, and in cases of disagreement unresolved by discussion, a third radiologist adjudicated to reach consensus.

## 2.5. Deep-learning Design & Feature Extraction

Multiple state-of-the-art DL architectures were systematically evaluated for their performance in ovarian lesion classification. The selected models represented different architectural approaches to medical image analysis, including traditional CNNs and modern ViT-based architectures (Table 1).

The CNN category included several proven architectures: (A) DenseNet variants including DenseNet121, DenseNet161, DenseNet169, and DenseNet201, which utilize dense connections between layers to improve feature propagation and reduce the vanishing gradient problem; (B) ResNet architectures including ResNet50, ResNet101, and ResNet152, employing residual connections to enable training of very deep networks; (C) EfficientNet models including EfficientNetB0 and EfficientNetB3, which optimize the balance between model accuracy and computational efficiency; (D) VGG architectures including VGG16 and VGG19, representing classical deep convolutional network designs; and (E) Xception architecture, which employs depthwise separable convolutions for improved parameter efficiency.

The transformer-based architectures included several variants of ViTs: ViT-Base with patch sizes of 16×16 and input resolutions of 224×224 (ViT16-224) and 384×384 (ViT16-384), and ViT-Base with patch sizes of 32×32 and input resolutions of 224×224 (ViT32-224) and 384×384 (ViT32-384).

Transfer learning strategies were employed to leverage pre-trained models and adapt them for ovarian lesion classification. All models were initially pre-trained on ImageNet, a large-scale natural image dataset containing over 14 million images across 1000 categories. This pre-training provides the models with fundamental visual feature recognition capabilities that can be adapted for medical image analysis. The transfer learning approach involved systematic unfreezing of network layers to optimize performance for the specific task of ovarian lesion classification. For CNNs, the final two to four layers were made trainable, while the earlier layers remained frozen, preserving low-level feature extraction capabilities and allowing task-specific adaptation. For ViT models, the final transformer encoder blocks were unfrozen for fine-tuning, enabling adaptation of attention mechanisms for medical image features (Table 1).

The training process was standardized across all models to ensure fair comparison. The configuration included a maximum of 100 training epochs with early stopping implemented when validation loss failed to improve for five consecutive epochs. The Adam optimizer was utilized with an initial learning rate of 0.003 and adaptive learning rate scheduling. A batch size of 16 was considered for each model architecture. Data augmentation techniques, including random resized crops (scale = (0.75, 1.00)), horizontal (factor = 0.5) and vertical (factor = 0.5) flipping, rotation up to 15 degrees, brightness and contrast adjustment, and random Gaussian blurring (kernel size = 5, sigma = (0.1, 1.5)) were applied to enhance model generalization.

A standardized preprocessing pipeline was implemented to ensure consistency across different acquisition devices and protocols. The preprocessing steps included DICOM to standard image format conversion while preserving metadata, intensity normalization using histogram equalization or z-score standardization, spatial resampling to consistent pixel dimensions of 224×224 or 384×384 pixels depending on model requirements, and noise reduction using spatial correlation filtering [13] when necessary.

### 2.6. Statistical Analysis & Model Performance Evaluation

The statistical computations were conducted using R statistical software (version 4.3.0). A robust cross-validation approach was implemented to ensure reliable performance estimation. The dataset was divided using stratified sampling to maintain a consistent representation of different lesion types and O-RADS categories across training and validation sets. A 4:1 ratio split was employed for the initial training-validation division. Evaluation was conducted using multiple metrics relevant to clinical decision-making, including area under the curve (AUC), sensitivity, specificity, and accuracy. In this study, a cut-off value of 3 in the O-RADS system was set to indicate malignancy and to evaluate the diagnostic performance of ACR O-RADS v2022. In addition, the combined prediction model integrating radiologist assessment and AI outputs was constructed using a binary logistic regression approach. Statistical significance was established at $p < 0.05$, and all confidence intervals were computed at the 95% confidence level. The Delong test was used for AUC comparison. All steps used in this study are depicted in Figure 1.

## 3. Results
### 3.1. Patient Characteristics
In total, 227 patients were included in this study, comprising 110 cases of at least one malignant ovarian cyst. A total of 512 ultrasound images derived from these patients were evaluated, with 259 images representing malignant cysts and 253 images representing benign cysts. The training cohort consisted of 182 patients (88 with malignant cysts and 94 with benign cysts), contributing 409 images (207 malignant, 202 benign). The validation cohort included 45 patients (22 malignant, 23 benign), yielding 103 images (52 malignant, 51 benign). The mean age was 40.7 years, ranging from 23 to 90 years.

### 3.2. O-RADS Interpretation by Radiologist
When evaluating ultrasound images according to the ACR O-RADS lexicon, the expert radiologist achieved an overall accuracy of 68.0% in differentiating malignancy from benign cysts in the validation set. The AUC for the radiologist's classification was 0.683, reflecting moderate discriminative performance (Table 2). The radiologist's sensitivity for identifying malignant cysts reached 38.5%, whereas the specificity for excluding benign cysts was 98.0%. Although the high specificity underscores the radiologist's conservative approach to categorizing benign cysts, the relatively low sensitivity highlights a propensity to undercall complex features, which may lead to missed opportunities for timely intervention.

### 3.3. Algorithm-Based Predictions
Most CNN and ViT architectures demonstrated superior overall discrimination compared with radiologist-only interpretation, with accuracy values ranging from 59.2% to 87.4% and AUCs spanning 0.620 to 0.941 (Table 2, Figure 2). Among the DenseNet family, DenseNet161 achieved the highest performance with an accuracy of 86.4% and an AUC of 0.898. DenseNet121 followed closely with an accuracy of 85.4% and an AUC of 0.894, while DenseNet169 recorded an accuracy of 81.6% and an AUC of 0.908. DenseNet201 yielded 80.6% accuracy and an AUC of 0.863.

EfficientNet variants attained moderate classification performance: EfficientNetB3 reached 68.0% accuracy with an AUC of 0.770, whereas EfficientNetB0 exhibited lower accuracy (59.2%) and an AUC of 0.700. Within the ResNet family, ResNet152 outperformed its counterparts with an accuracy of 81.5% and an AUC of 0.877. ResNet50 and ResNet101 achieved accuracies of 62.1% and 61.2% with AUCs of 0.698 and 0.620, respectively.

Classic CNNs such as VGG16 and VGG19 also demonstrated strong discriminative capabilities, with VGG16 attaining 79.6% accuracy and an AUC of 0.866, and VGG19 yielding 73.8% accuracy and an AUC of 0.806. The Xception model produced 72.8% accuracy with an AUC of 0.858.

ViT architectures delivered the highest metrics among all models. ViT16-224 achieved 86.4% accuracy and an AUC of 0.912, while ViT32-224 attained 80.6% accuracy and an AUC of 0.857. Larger input sizes further enhanced performance: ViT32-384 produced 84.5% accuracy and an AUC of 0.922, and ViT16-384 achieved the greatest overall discrimination with 87.4% accuracy and an AUC of 0.941 (Table 2, Figure 2).

*3.4. Synergistic Human-Machine Diagnostic Performance*

When radiologist-derived O-RADS scores were fused with DL outputs, all model families showed further gains in accuracy and discrimination, although the statistical significance of these improvements varied across architectures (Table 2, Figure 2). In the DenseNet family, the standalone DenseNet121 achieved 85.4% accuracy with an AUC of 0.894, and when combined with O-RADS, this rose to 88.3% and an AUC of 0.919 (P = 0.452). DenseNet161 alone reached 86.4% accuracy (AUC 0.898), which improved to 90.3% and 0.934 upon integration of O-RADS (P = 0.024). The DenseNet169 exhibited 81.6% accuracy (AUC=0.908) that increased to 88.3% (AUC=0.935) with O-RADS (P = 0.145). DenseNet201's standalone metrics were 80.6% accuracy and AUC 0.863, improving to 88.3% accuracy and 0.920 AUC in the hybrid model (P = 0.034).

Among EfficientNet architectures, EfficientNet_B0 achieved 59.2% accuracy (AUC 0.700), which rose to 77.7% and 0.808 with O-RADS (P = 0.002). EfficientNet_B3 increased from 68.0% accuracy (AUC 0.770) to 82.5% (AUC 0.855) when combined with O-RADS (P = 0.013). In the ResNet series, ResNet50 improved from 62.1% accuracy and AUC 0.698 to 80.6% and 0.813 with O-RADS (P = 0.003). ResNet101 rose from 61.2% and 0.620 to 72.8% and 0.744 with O-RADS (P = 0.002). ResNet152's standalone performance of 81.5% accuracy and AUC 0.877 improved to 86.4% and 0.903 in the hybrid framework (P = 0.099).

Classic CNN models saw comparable and significant gains (P<0.05): VGG16+O-RADS achieved 86.4% accuracy and an AUC of 0.931, VGG19+O-RADS reached 84.5% accuracy with an AUC of 0.886, and Xception+O-RADS attained 88.3% accuracy and an AUC of 0.898.

Hybrid human–AI combinations with ViTs displayed the highest absolute performance. ViT32-224 yielded 80.6% accuracy (AUC=0.857), which improved to 86.4% and 0.897 with O-RADS (P = 0.102). ViT16-224 rose from 86.4% (AUC=0.912) to 91.3% (AUC=0.942) (P = 0.185). ViT32-384 increased from 84.5% accuracy (AUC 0.922) to 90.3% (AUC 0.934) (P = 0.415). Finally, the top performer, ViT16-384, advanced from 87.4% accuracy (AUC 0.941) to 91.3% (AUC 0.952) when combined with O-RADS (P = 0.455). These findings demonstrate that human–machine fusion enhances diagnostic performance across all model families, with the most significant improvements observed in DenseNet161 and ResNet50 hybrids. Representative clinical examples illustrating the comparative diagnostic performance across radiologist-only, AI-only, and combined approaches are provided in Figures 3 and 4.

## 4. Discussion

The present study provides the first comprehensive comparison of radiologist performance using the ACR O-RADS Ultrasound v2022 classification against a suite of contemporary DL architectures, as well as the synergistic performance achieved by integrating radiologist-derived O-RADS scores with AI outputs. The findings demonstrate that while expert radiologists applying ACR O-RADS v2022 exhibit moderate discriminative performance in identifying malignant ovarian lesions, state-of-the-art DL models substantially outperform human assessment alone. The combination of human expertise and computational analysis yields the highest diagnostic accuracy and discrimination. These results have important implications for clinical practice, suggesting that AI assistance can meaningfully enhance the sensitivity and specificity of ovarian lesion stratification, ultimately improving patient outcomes.

Despite its advancements over earlier iterations, ACR O-RADS v2022 retains inherent limitations when used in isolation, particularly in its sensitivity for detecting complex malignant features. In our validation cohort, the expert radiologist attained an overall accuracy of 68.0% and an AUC of 0.683, with a pronounced disparity between sensitivity (38.5%) and specificity (98.0%). This pattern reflects a conservative approach that minimizes false-positive categorization of simple cysts but risks undercalling malignant or complex lesions, potentially delaying timely intervention. Comparable single-center and multi-institutional evaluations of ACR O-RADS v2022 have reported AUC values ranging from 0.871 to 0.940 and sensitivities between 90.6% and 96.6%, with specificities of 81.9% to 98.0%, depending on patient selection and prevalence of malignancy [14, 15]. Notably, in nonselected cohorts representing routine clinical practice, diminished prevalence of cancer correlates with lower specificity and positive predictive value, underscoring the challenge of applying ACR O-RADS v2022 across diverse populations.

In contrast, DL models evaluated in this study exhibited superior standalone performance. CNNs such as DenseNet161 achieved an AUC of 0.898 and an accuracy of 86.4%, while classic architectures like VGG16 and Xception yielded AUCs of 0.866 and 0.858, respectively. ViT variants further advanced discrimination, with ViT16-384 reaching an AUC of 0.941 and an accuracy of 87.4%. These findings align with earlier investigations demonstrating that AI algorithms can match or exceed expert radiologists in ovarian lesion classification and supporting their potential as robust decision-support tools [16]. The superior performance of transformer-based architectures likely reflects their capacity to model long-range spatial dependencies and attend to salient image regions, augmenting feature representation beyond the localized receptive fields of traditional CNNs.

Importantly, the integration of radiologist-derived O-RADS scores with AI outputs consistently enhanced diagnostic metrics across all model families. Human-machine fusion approaches, implemented here via logistic regression models combining O-RADS categorical scores with DL-predicted probabilities, yielded the highest performance. These gains reflect the complementary strengths of human expertise in contextual pattern recognition and AI's sensitivity to subtle image cues. Prior studies evaluating hybrid approaches with earlier O-RADS versions have reported similar enhancements, with combined AUCs increasing by up to 0.05 and accuracy gains of 5-10% over AI-only or radiologist-only methods [11, 17]. The current work extends these observations to the latest ACR O-RADS v2022 system and a broader array of AI architectures, confirming that expert guidance remains indispensable even as AI capabilities advance. On the other hand, comparing the results of CNNs and ViTs revealed that human–machine fusion approaches could significantly improve CNNs, while no significant improvement was observed for ViTs. This finding suggests that the ViT architecture can extract richer and more powerful patterns from ovarian ultrasound images, and that O-RADS could not provide any additional value to them.

Beyond quantitative performance, the hybrid human-AI paradigm offers practical clinical advantages [18, 19]. Enhanced specificity reduces false positives and unnecessary surgical interventions, while elevated sensitivity ensures that high-risk lesions receive prompt evaluation. In settings where less experienced sonographers or non-subspecialty radiologists interpret pelvic ultrasound studies, AI assistance could standardize risk stratification, mitigate variability, and improve diagnostic confidence [20]. Moreover, by highlighting discrepant cases, where AI and human assessments divergent decision-support systems can prompt secondary reviews or additional imaging, fostering more efficient resource utilization [21].

Future research should prioritize prospective, multicenter trials to assess the real-world impact of AI-augmented ACR O-RADS v2022 on clinical outcomes, including diagnostic turnaround times, referral patterns, and patient-centered endpoints. Integrating complementary data streams, including serum biomarkers (e.g., CA125, HE4), contrast-enhanced ultrasound parameters, and Doppler flow metrics into multimodal fusion models may further refine lesion characterization and risk stratification. In addition, cost-effectiveness analyses and workflow integration studies will ultimately determine the feasibility of widespread adoption and inform reimbursement strategies for hybrid human-AI diagnostic paradigms.

Despite these findings, several clinical and methodological limitations warrant consideration. First, the single-center, retrospective design may limit generalizability to broader patient populations. Second, reliance on static two-dimensional images (as O-RADS v2022 categories rely on static lexicons) rather than real-time cine loops may underrepresent subtle lesion features and requires further investigation. Finally, the time requirements for AI processing and report integration were not measured and compared with those of a radiologist alone, limiting workflow feasibility conclusions.

## 5. Conclusion

In summary, AI models significantly outperformed radiologist-only ACR O-RADS v2022 assessment in differentiating benign from malignant ovarian lesions. The integration of radiologist-derived O-RADS scores with AI predictions yielded the highest diagnostic accuracy and discrimination. Hybrid human–AI frameworks promise to enhance clinical decision-making by reducing false positives and improving sensitivity for high-risk lesions. Prospective multicenter validation and robust explainability measures are needed to ensure generalizability and clinician trust.

**Table 1.** Specifications of the models and their layers utilized for transfer learning in this study.

| Type | Model | Input Size | Patch Size | Trainable Layers |
|---|---|---|---|---|
| **Convolutional Neural Network** | DenseNet121 | 224×224×3 | - | DenseBlock3, DenseBlock4 |
| | DenseNet161 | 224×224×3 | - | DenseBlock3, DenseBlock4 |
| | DenseNet169 | 224×224×3 | - | DenseBlock3, DenseBlock4 |
| | DenseNet201 | 224×224×3 | - | DenseBlock3, DenseBlock4 |
| | Efficientnet_B0 | 224×224×3 | - | Block 7, Block 8 |
| | Efficientnet_B3 | 300×300×3 | - | Block 7, Block 8 |
| | ResNet50 | 224×224×3 | - | Layer3, layer4 |
| | ResNet101 | 224×224×3 | - | Layer3, layer4 |
| | ResNet152 | 224×224×3 | - | Layer3, layer4 |
| | VGG16 | 224×224×3 | - | Last two convolutional layers |
| | VGG19 | 224×224×3 | - | Last two convolutional layers |
| | Xception | 299×299×3 | - | Block11, Block12 |
| **Vision Transformer** | Vit_base_patch16_384 | 384×384×3 | 16×16×3 | Last two Transformer encoder blocks |
| | Vit_large_patch16_224 | 224×224×3 | 16×16×3 | Last two Transformer encoder blocks |
| | Vit_base_patch32_224 | 224×224×3 | 32×32×3 | Last two Transformer encoder blocks |
| | Vit- base -patch32-384 | 384×384×3 | 32×32×3 | Last two Transformer encoder blocks |

**Table 2.** Diagnostic performance of deep-learning models based on the validation dataset in classifying benign and malignant ovarian cysts in ultrasound images.

| Method | Predicted Class | True Class | | Kappa with Reference Standard | Sen (%) | Spec (%) | Acc (%) | AUC (95% CI) | P-value |
|---|---|---|---|---|---|---|---|---|---|
| | | Malignant | Benign | | | | | | |
| DenseNet121 | Malignant | 45 | 8 | 0.709 | 86.6 | 84.3 | 85.4 | 0.894 (0.830, 0.958) | 0.452 |
| | Benign | 7 | 43 | | | | | | |
| DenseNet121+O-RADS | Malignant | 45 | 5 | 0.767 | 86.6 | 90.2 | 88.3 | 0.919 (0.861, 0.978) | |
| | Benign | 7 | 46 | | | | | | |
| DenseNet161 | Malignant | 46 | 8 | 0.728 | 88.5 | 84.3 | 86.4 | 0.898 (0.831, 0.965) | **0.024** |
| | Benign | 6 | 43 | | | | | | |
| DenseNet161+O-RADS | Malignant | 48 | 6 | 0.806 | 92.3 | 88.2 | 90.3 | 0.934 (0.880, 0.988) | |
| | Benign | 4 | 45 | | | | | | |
| DenseNet169 | Malignant | 44 | 11 | 0.631 | 84.6 | 78.4 | 81.6 | 0.908 (0.853, 0.963) | 0.145 |
| | Benign | 8 | 40 | | | | | | |
| DenseNet169+O-RADS | Malignant | 45 | 5 | 0.767 | 86.5 | 90.2 | 88.3 | 0.935 (0.888, 0.982) | |
| | Benign | 7 | 46 | | | | | | |
| DenseNet201 | Malignant | 39 | 7 | 0.612 | 75.0 | 86.3 | 80.6 | 0.863 (0.787, 0.940) | **0.034** |
| | Benign | 13 | 44 | | | | | | |
| DenseNet201+O-RADS | Malignant | 47 | 7 | 0.767 | 90.4 | 86.3 | 88.3 | 0.920 (0.861, 0.980) | |
| | Benign | 5 | 44 | | | | | | |
| EfficientNet_B0 | Malignant | 42 | 32 | 0.181 | 80.8 | 37.2 | 59.2 | 0.700 (0.597, 0.803) | **0.002** |
| | Benign | 10 | 19 | | | | | | |
| EfficientNet_B0+O-RADS | Malignant | 39 | 10 | 0.554 | 75.0 | 80.4 | 77.7 | 0.808 (0.721, 0.896) | |
| | Benign | 13 | 41 | | | | | | |
| EfficientNet_B3 | Malignant | 40 | 21 | 0.358 | 76.9 | 58.8 | 68.0 | 0.770 (0.680, 0.861) | **0.013** |
| | Benign | 12 | 30 | | | | | | |
| EfficientNet_B3+O-RADS | Malignant | 36 | 2 | 0.651 | 69.2 | 96.1 | 82.5 | 0.855 (0.781, 0.929) | |
| | Benign | 16 | 49 | | | | | | |
| ResNet50 | Malignant | 38 | 25 | 0.241 | 73.1 | 51.0 | 62.1 | 0.698 (0.593, 0.803) | **0.003** |
| | Benign | 14 | 26 | | | | | | |
| ResNet50+O-RADS | Malignant | 42 | 10 | 0.612 | 80.8 | 80.4 | 80.6 | 0.813 (0.729, 0.898) | |
| | Benign | 10 | 41 | | | | | | |
| ResNet101 | Malignant | 34 | 22 | 0.223 | 65.4 | 56.9 | 61.2 | 0.620 (0.511, 0.729) | **0.002** |
| | Benign | 18 | 29 | | | | | | |
| ResNet101+O-RADS | Malignant | 26 | 2 | 0.459 | 50.0 | 96.1 | 72.8 | | |
| | Benign | 26 | 49 | | | | | 0.744 (0.649, 0.840) | |
| **ResNet152** | Malignant | 41 | 8 | 0.631 | 78.8 | 84.3 | 81.5 | 0.877 (0.809, 0.946) | 0.099 |
| | Benign | 11 | 43 | | | | | | |
| **ResNet152+O-RADS** | Malignant | 46 | 8 | 0.728 | 88.5 | 84.3 | 86.4 | 0.903 (0.840, 0.967) | |
| | Benign | 6 | 43 | | | | | | |
| **VGG16** | Malignant | 44 | 13 | 0.592 | 84.6 | 74.5 | 79.6 | 0.866 (0.796, 0.936) | **0.019** |
| | Benign | 8 | 38 | | | | | | |
| **VGG16+O-RADS** | Malignant | 44 | 6 | 0.728 | 84.6 | 88.2 | 86.4 | 0.931 (0.886, 0.976) | |
| | Benign | 8 | 45 | | | | | | |
| **VGG19** | Malignant | 33 | 8 | 0.477 | 63.5 | 84.3 | 73.8 | 0.806 (0.721, 0.890) | **0.019** |
| | Benign | 19 | 43 | | | | | | |
| **VGG19+O-RADS** | Malignant | 45 | 9 | 0.689 | 86.5 | 82.3 | 84.5 | 0.886 (0.823, 0.948) | |
| | Benign | 7 | 42 | | | | | | |
| **Xception** | Malignant | 44 | 20 | 0.453 | 84.6 | 60.8 | 72.8 | 0.858 (0.781, 0.935) | **0.045** |
| | Benign | 8 | 31 | | | | | | |
| **Xception+O-RADS** | Malignant | 43 | 3 | 0.767 | 82.7 | 94.1 | 88.3 | 0.898 (0.832, 0.964) | |
| | Benign | 9 | 48 | | | | | | |
| **ViT32-224** | Malignant | 40 | 8 | 0.612 | 76.9 | 84.3 | 80.6 | 0.857 (0.784, 0.931) | 0.102 |
| | Benign | 12 | 43 | | | | | | |
| **ViT32-224+O-RADS** | Malignant | 43 | 5 | 0.728 | 82.7 | 90.2 | 86.4 | 0.897 (0.834, 0.960) | |
| | Benign | 9 | 46 | | | | | | |
| **ViT16-224** | Malignant | 42 | 4 | 0.728 | 80.8 | 92.2 | 86.4 | 0.912 (0.856, 0.969) | 0.185 |
| | Benign | 10 | 47 | | | | | | |
| **ViT16-224+O-RADS** | Malignant | 47 | 4 | 0.825 | 90.4 | 92.2 | 91.3 | 0.942 (0.894, 0.989) | |
| | Benign | 5 | 47 | | | | | | |
| **ViT32-384** | Malignant | 48 | 12 | 0.689 | 92.3 | 76.5 | 84.5 | 0.922 (0.869, 0.975) | 0.415 |
| | Benign | 4 | 39 | | | | | | |
| **ViT32-384+O-RADS** | Malignant | 49 | 7 | 0.806 | 94.2 | 86.3 | 90.3 | 0.934 (0.884, 0.984) | |
| | Benign | 3 | 44 | | | | | | |
| **ViT16-384** | Malignant | 43 | 4 | 0.728 | 82.7 | 92.2 | 87.4 | 0.941 (0.897, 0.984) | 0.455 |
| | Benign | 9 | 47 | | | | | | |
| **ViT16-384+O-RADS** | Malignant | 47 | 4 | 0.825 | 90.4 | 92.2 | 91.3 | 0.952 (0.910, 0.994) | |
| | Benign | 5 | 47 | | | | | | |
| **O-RADS** | Malignant | 20 | 1 | 0.363 | 38.5 | 98.0 | 68.0 | 0.683 (0.578, 0.787) | |
| | Benign | 32 | 50 | | | | | | |

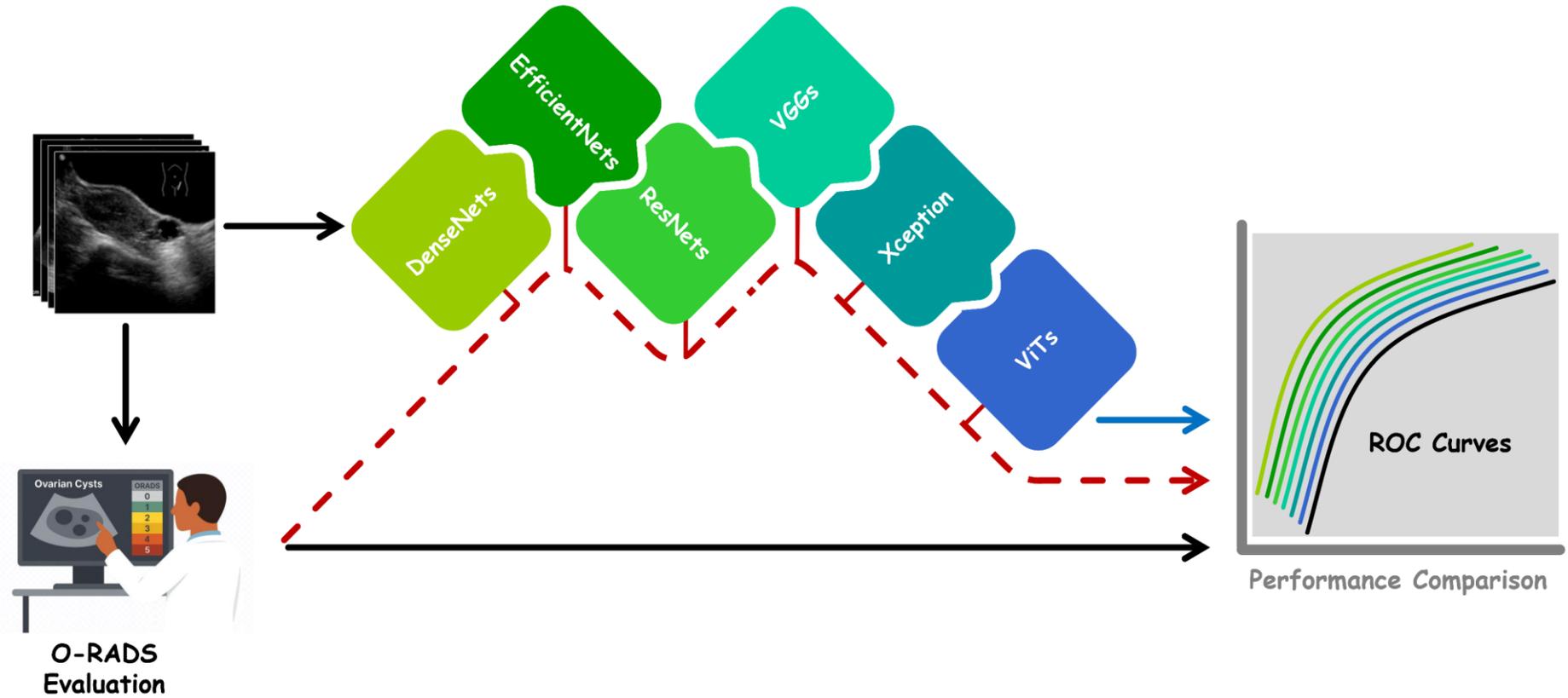

**Fig. 1.** Overview of the steps used in this study. The performance of each deep-learning model was evaluated with and without ACR O-RADS to highlight its contribution.

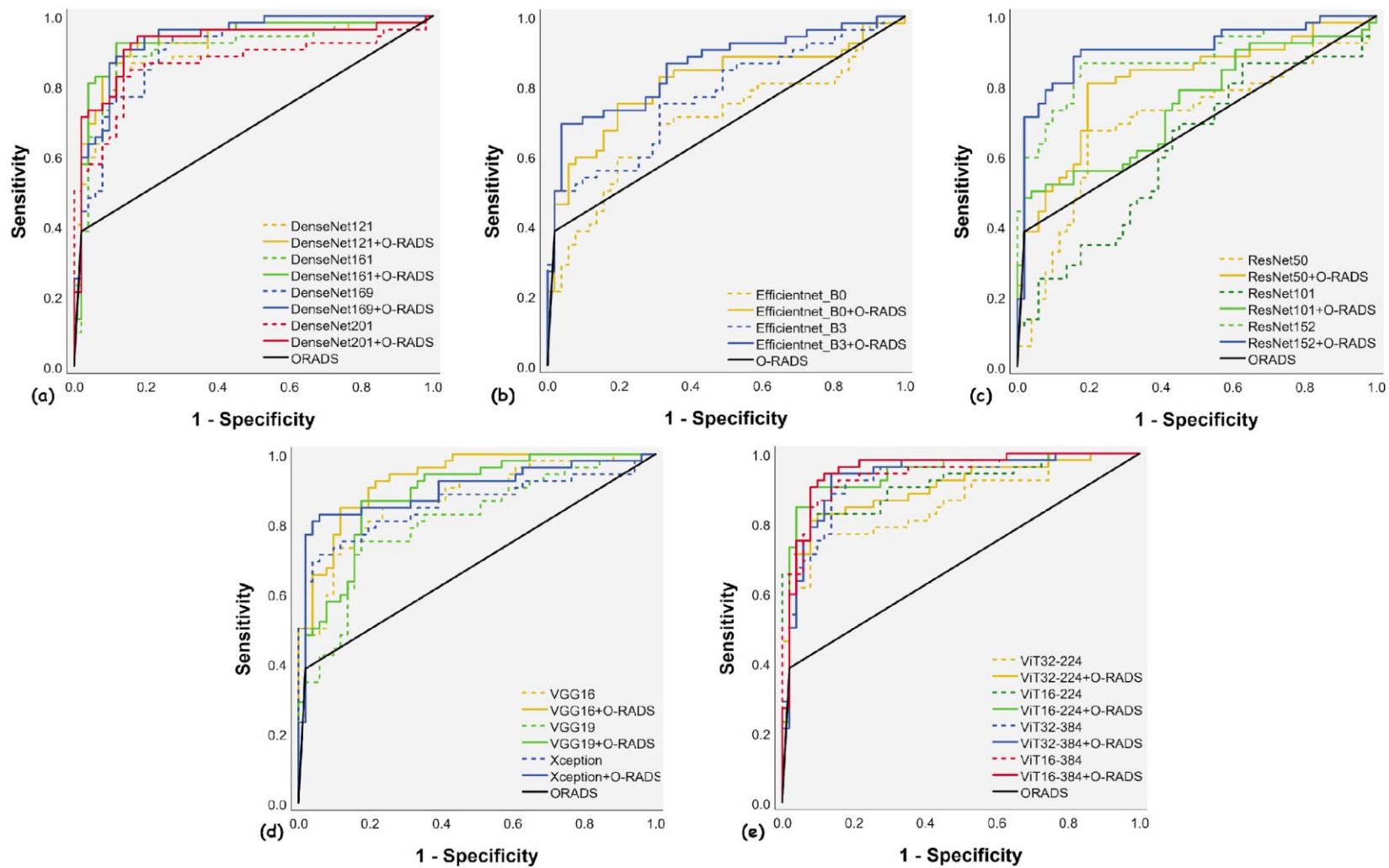

**Fig. 2.** ROC curves of sixteen deep learning models with and without O-RADS scores: (a) DenseNet; (b) EfficientNet; (c) ResNet; (d) VGG and Xception; and (e) Vision Transformer families. The ROC curve of O-RADS is included in all sub-sections for better comparison.

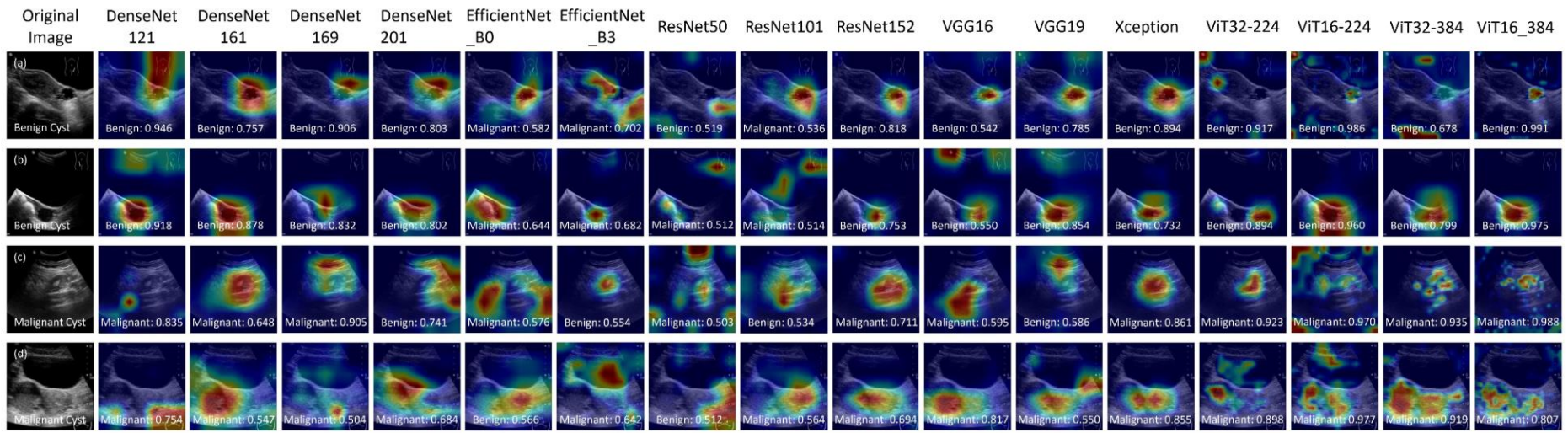

**Fig. 3.** Sample ultrasound images of ovarian cysts with their corresponding Grad-CAM visualizations, labels, and predicted probabilities of sixteen deep learning models. (a–b) Benign ovarian cysts, correctly assigned an O-RADS score of 2 by the radiologist. (c) A malignant ovarian cyst, correctly assigned an O-RADS score of 3. (d) A malignant ovarian cyst, initially misclassified by the radiologist with an O-RADS score of 2. The AI-assisted model helped the radiologist correct the scoring.

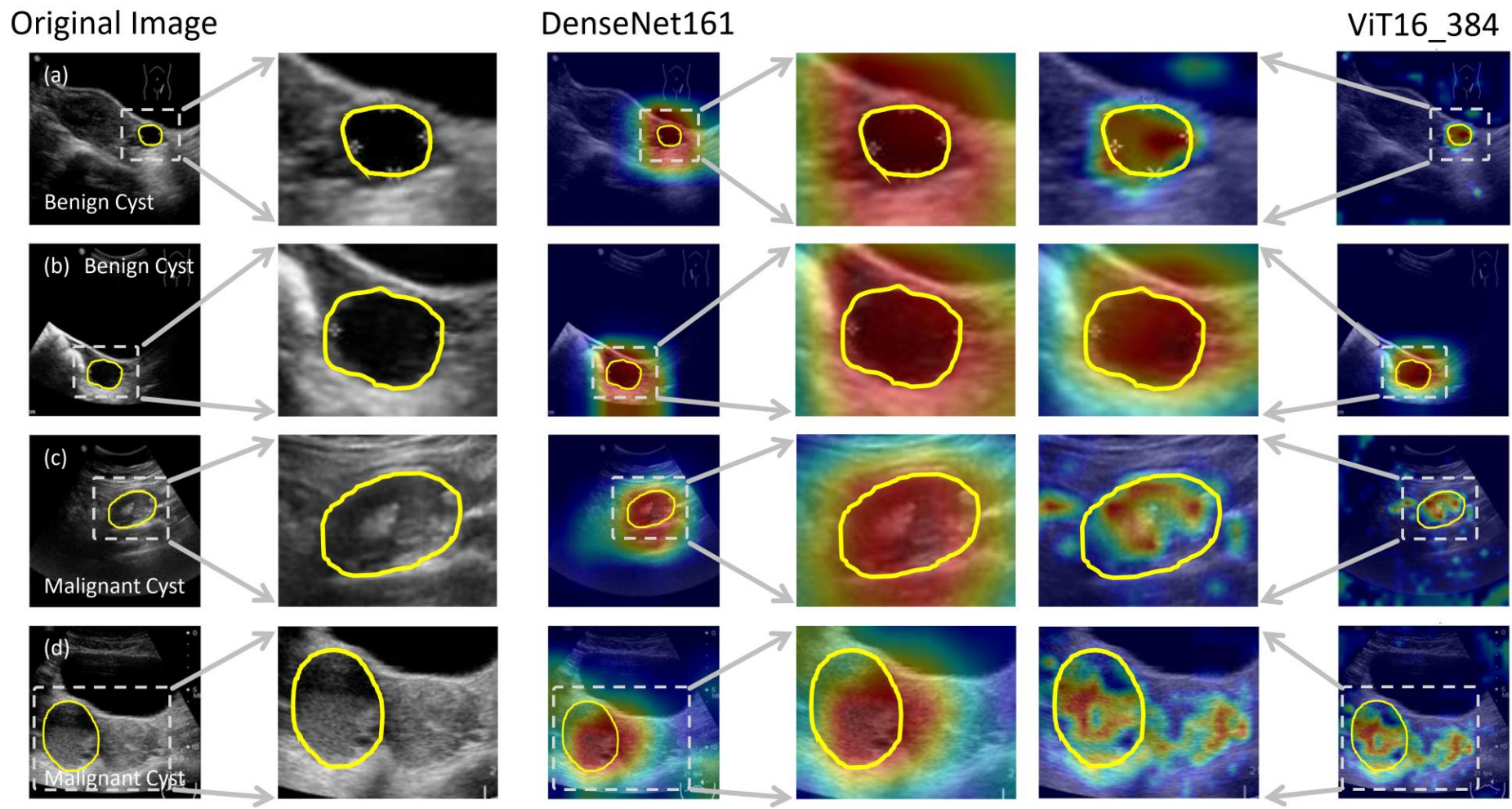

**Fig. 4.** Sample ultrasound images of ovarian cysts with their corresponding Grad-CAM visualizations from the best CNN model, DenseNet161, and the best ViT model, ViT16-384, shown with zoomed-in views for clearer comparison.